\documentclass{article}

\usepackage{arxiv}

\usepackage[utf8]{inputenc} 
\usepackage[T1]{fontenc}    
\usepackage{hyperref}       
\usepackage{url}            
\usepackage{booktabs}       
\usepackage{amsfonts}       
\usepackage{graphicx}
\usepackage{doi}
\def\BibTeX{{\rm B\kern-.05em{\sc i\kern-.025em b}\kern-.08em
    T\kern-.1667em\lower.7ex\hbox{E}\kern-.125emX}}

\title{Fast and Real-time End to End Control in
Autonomous Racing Cars Through Representation
Learning}

\author{{\hspace{1mm}Praveen Venkatesh} \\
	Department of Electrical Engineering\\
	Indian Institute of Technology\\
	Gandhinagar, India \\
	\texttt{praveen.venkatesh@iitgn.ac.in} \\
	\And
	{\hspace{1mm}Rwik Rana} \\
	Department of Mechanical Engineering\\
	Indian Institute of Technology\\
	Gandhinagar, India \\
	\texttt{rwik.rana@iitgn.ac.in} \\
	\And
	{\hspace{1mm}Harish PM}\\
	Department of Mechanical Engineering\\
	Indian Institute of Technology\\
	Gandhinagar, India  \\
	\texttt{harish@iitgn.ac.in} \\
}

\date{}

\renewcommand{\shorttitle}{Autonomous Race Car}

\begin{document}
\maketitle


\begin{abstract}
The challenges presented in an autonomous racing situation are distinct from those faced in regular autonomous driving and require faster end-to-end algorithms and consideration of a longer horizon in determining optimal current actions keeping in mind upcoming maneuvers and situations. In this paper, we propose an end-to-end method for autonomous racing that takes in as inputs video information from an onboard camera and determines final steering and throttle control actions. We use the following split to construct such a method (1) learning a low dimensional representation of the scene, (2) pre-generating the optimal trajectory for the given scene, and (3) tracking the predicted trajectory using a classical control method. In learning a low-dimensional representation of the scene, we use intermediate representations with a novel unsupervised trajectory planner to generate expert trajectories, and hence utilize them to directly predict race lines from a given front-facing input image. Thus, the proposed algorithm employs the best of two worlds - the robustness of learning-based approaches to perception and the accuracy of optimization-based approaches for trajectory generation in an end-to-end learning-based framework. We deploy and demonstrate our framework on CARLA, a photorealistic simulator for testing self-driving cars in realistic environments.
   
\end{abstract}

\keywords{Autonomous Racing, End-to-end control, Representation Learning, Reinforcement Learning} 

\section{Introduction}
The goal of a racecar is to complete a given track in the least possible time. Over several decades of motorsport racing events, the racing speeds have increased tremendously, with car speeds reaching up to up to 360 Km/h in some races. As a consequence of such high racing speeds, the time available for decision making in a race car is vastly diminished. Furthermore, the decision-making is also more complex than a regular lane-driving situation and involves race drivers having to modulate their positioning within the track width and their speeds optimally in anticipation of upcoming maneuvers (sometimes several hundred meters ahead of time). Thus the racing situation distinguishes itself from a regular lane driving situation in terms of three key challenges; less time for decision-making, higher importance of controlling positioning within the track width as compared to lane driving, and the necessity to consider longer future horizon (e.g. upcoming maneuvers) in decision-making \cite{7963748,8754713,morari15}. In response to the latter two challenges above, a racecar driver attempts to maintain a racing line that takes into account speed and upcoming maneuver to position the car within the track width to optimize the time taken to cover the track \cite{TALVALA2011137}.

In this work, we consider control for an autonomous racecar driving situation. The motivation for considering an autonomous racing situation is twofold. One, the set of challenges presented in a racing situation is distinct from a regular driving situation, as highlighted earlier. Further, in an autonomous situation, a fourth challenge is to substitute the racecar driver's perception of the track and maneuver ahead with an autonomous algorithm that can run fast enough to be useful in a real-time race driving situation. Two, since a racing situation forces the control strategy to consider a longer future horizon in the decision-making, the developed solutions will be useful in other control situations requiring longer horizon consideration.


Autonomous racing has been explored in various previous works \cite{7963748,8754713,morari15} and the autonomous tasks are traditionally bifurcated into three components - (1) Perception \& State estimation, (2) Motion Planning, and (3) Control. Several works have explored replacing each of these systems with learning-based approaches. Deep learning methods for perception have shown tremendous success in vision-based applications. Compared to classical vision-based approaches, learning-based methods are robust to errors and are often faster. Traditional methods that determine the racing line utilize optimization strategies that rely on robust state estimation and mapping and are often computationally expensive. Furthermore, these strategies rely on accurate models of the car's dynamics, leading to a trade-off between accuracy and computational requirements. Likewise, substantial literature exists on motion planning and control in various driving situations, but the literature on motion planning and control for autonomous racing situation is still relatively in its infancy.

In our work, we propose an end-to-end method for autonomous racing that takes in as inputs video information from an on-board camera and determines final steering and throttle control actions. We use the following split to construct such a method (1) learning a low dimensional representation of the scene, (2) pre-generating the optimal trajectory for the given scene, and (3) tracking the predicted trajectory using a classical control method. In learning a low-dimensional representation of the scene, we use intermediate representations with a novel unsupervised expert trajectory generation technique to directly predict race lines from a given front-facing input image. Based on this data set generated, we employ a supervised learning algorithm to learn an end-to-end control strategy. Thus, the proposed algorithm employs the best of two worlds - the robustness of learning-based approaches to perception and the accuracy of optimization-based approaches for trajectory generation in an end-to-end learning-based framework. 

We deploy our framework on CARLA, a photo-realistic simulator for testing self-driving cars in realistic environments. CARLA supports realistic dynamics of vehicles allowing us to simulate and verify our proposed framework. We develop this framework with an intention to deploy on a hardware framework currently being developed in the lab (with a camera with depth sensing and with fully on-board computation), but leave the hardware implementation as part of future work.

\section{Related Approaches}

Prior attempts at developing controls for autonomous racing may be loosely categorized into two sets of approaches, classical approaches and learning-based approaches.

\textbf{Classical Approach}: The crudest of classical methods are based on purely reactive (to lane markers or obstacles) decision-making that does not incorporate trajectory planning and does not take into account future predictions. Other approaches build on a SLAM framework \cite{lienkamp2019} for perception and a separate trajectory planner. Trajectory planning or generation often involves accurate state estimation obtained in the form of a bird's eye view of the environment (Cartesian space) \cite{loukkal_driving_2021}. Alternatively, trajectories are generated using search-based methods such as graph search \cite{https://doi.org/10.1002/atr.1359}, or incremental methods such as RRT and SST \cite{li_sparse_2015}. The third set of approaches focus on an optimization framework or model predictive control that takes into account a prediction horizon in the decision making  \cite{ALCALA2020104270,7963748,8754713,morari15}. These methods have shown promise, but some of these approaches are computationally expensive and sometimes are inadequate for the agile sport of racing.


\textbf{Learning based Approach}: Learning-based approaches use two distinct approaches - supervised and unsupervised. Among supervised learning methods, \cite{DBLP:journals/corr/abs-2010-08776} propose a framework to regress directly from images to steering angle at a fixed speed. However, these do not take any time-dependent situations into account and may impact performance. AdmiralNet \cite{9116486} attempts to solve this problem by introducing LSTM cells to enforce time dependence. Other learning-based methods aim to predict the required waypoints to be tracked instead of the steering angle directly \cite{altche_lstm_2017}. This approach can suffer from the drawback that if there is a deviation or inaccuracy in a single waypoint, it may lead to non-smooth trajectories and subsequent degradation in performance at high speeds. \cite{weiss_bezier_nodate} solve this problem by utilizing a method to regress to waypoints based on Bezier curve interpolations leading to smooth trajectories. However, supervised methods are limited by the expert demonstrations provided to the network and cannot achieve performance better than the demonstrations themselves.

Unsupervised learning methods, on the other hand, try to solve this problem by using reinforcement learning-based techniques. \cite{e2erl} uses an end to end RL framework to learn to drive a racecar through a variety of different terrain, while \cite{superhuman} use a mixture of high and low dimensional features to achieve superhuman performance in a racing game. \cite{flatmobiles} utilize a method for predicting the OGM of a given image and employ waypoint prediction using an autoencoder. This method poses several advantages as it is significantly easier to plan trajectories from an accurate estimation of the occupancy grid.

The proposed method in this paper exploits the ease of planning using an occupancy grid map used by traditional methods and the advantages of unsupervised learning for obtaining trajectories that outperform human drivers.

\section{Proposed Method}

The proposed pipeline consists of the following stages - (1) Learning a low dimensional representation of the scene, (2) Generating the optimal trajectory for the given scene. (3) Tracking the desired trajectory using a classical control method. We write our experiments on PyTorch \cite{NEURIPS2019_9015} and run our simulation on CARLA \cite{Dosovitskiy17}.
    
\subsection{Perception}

\begin{figure}
    \centering
    \includegraphics[width = \linewidth]{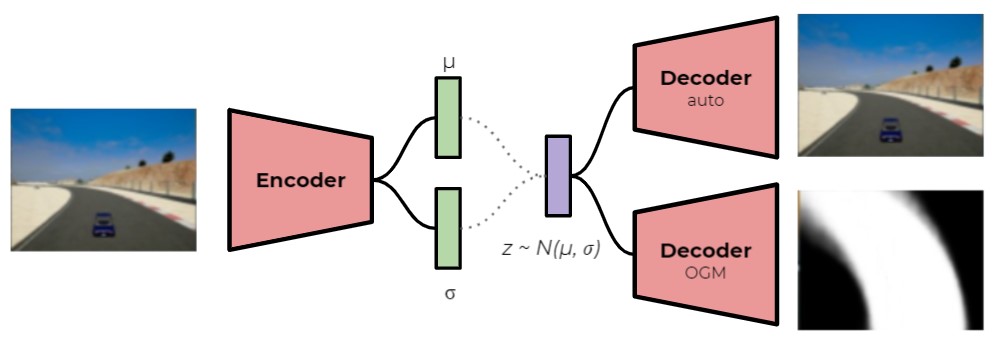}
    \caption{Perception Block - Cross Modal VAE architecture to obtain a compressed latent representation of the environment with respect to the bird's eye view } 
    \label{fig:cmvae}
\end{figure}

VAE's have been used widely for generative modelling of a given distribution of data. Since our primary goal is to learn a low dimensional representation that can represent both the image space as well as the object graph mapper (OGM), we utilize a cross-modal approach to training the VAE, utilizing both the input image that needs to be autoencoded, as well as the OGM of the environment. \cite{airsim} show that the CMVAE architecture is robust to the simulator-reality domain transfer and also implicitly regularizes the learnt latent space representation. A block diagram of our architecture can be seen in Fig \ref{fig:cmvae}.

Our architecture primarily attempts to learn a low dimensional representation of the lanes and any surrounding obstacles to predict trajectories.


 
The input to our framework is a monocular front-facing RGB image ($\mathbb{R} ^3$). We model our problem as a mapping from $\mathbb{R} ^{3\times H\times W} \rightarrow \mathbb{R}^{k}$, where the output is a $k$ dimensional embedding that represents a set of control points for trajectory prediction.

\subsection{Trajectory Planning}

To generate the training data required for trajectory estimation, we use a novel approach based on simplifying the environment. We use the occupancy maps generated from the top views to set up a simulation and use a learning-based speciation strategy to learn optimal trajectories that the car has to follow. 

We represent the trajectories as a set of $(x, y)$ coordinates generated by the trajectory planner. The trajectories are approximated with a single Bezier curve, where points are resampled at regular $y$ intervals. The $x$ coordinates of the samples form the \textit{trajectory embedding} which is meant to be the output representing the trajectory for the car to follow using a traditional controller.

\subsubsection{Vehicle Environment}
A simplified custom OGM environment is created to imitate the top view of the road. The environment supports multiple agents that can be spawned on a single track, with each agents' movement treated independently. The goal of this environment is to train a lightweight feedforward neural network to generate a trajectory given the OGM of the car and the road at every instant. We use a simple 3 layer multi-layer perceptron with $\lambda$ distance inputs for the agent that learns to generate trajectories.

\textbf{Agent Modelling} : We use a bicycle model to represent the simplified car for trajectory generation. The bicycle model takes into consideration the yaw angle, heading angle, the direction of velocity and friction with the ground. The parameters $l_r$, $l_f$, mass (M) of the vehicle was kept considering Hobbywing EZRUN Max10 60A as our test car, on which the final system would eventually be deployed.

\begin{figure}[h]
\centering
    \begin{minipage}[b]{0.45\linewidth}
        \includegraphics[scale =0.05]{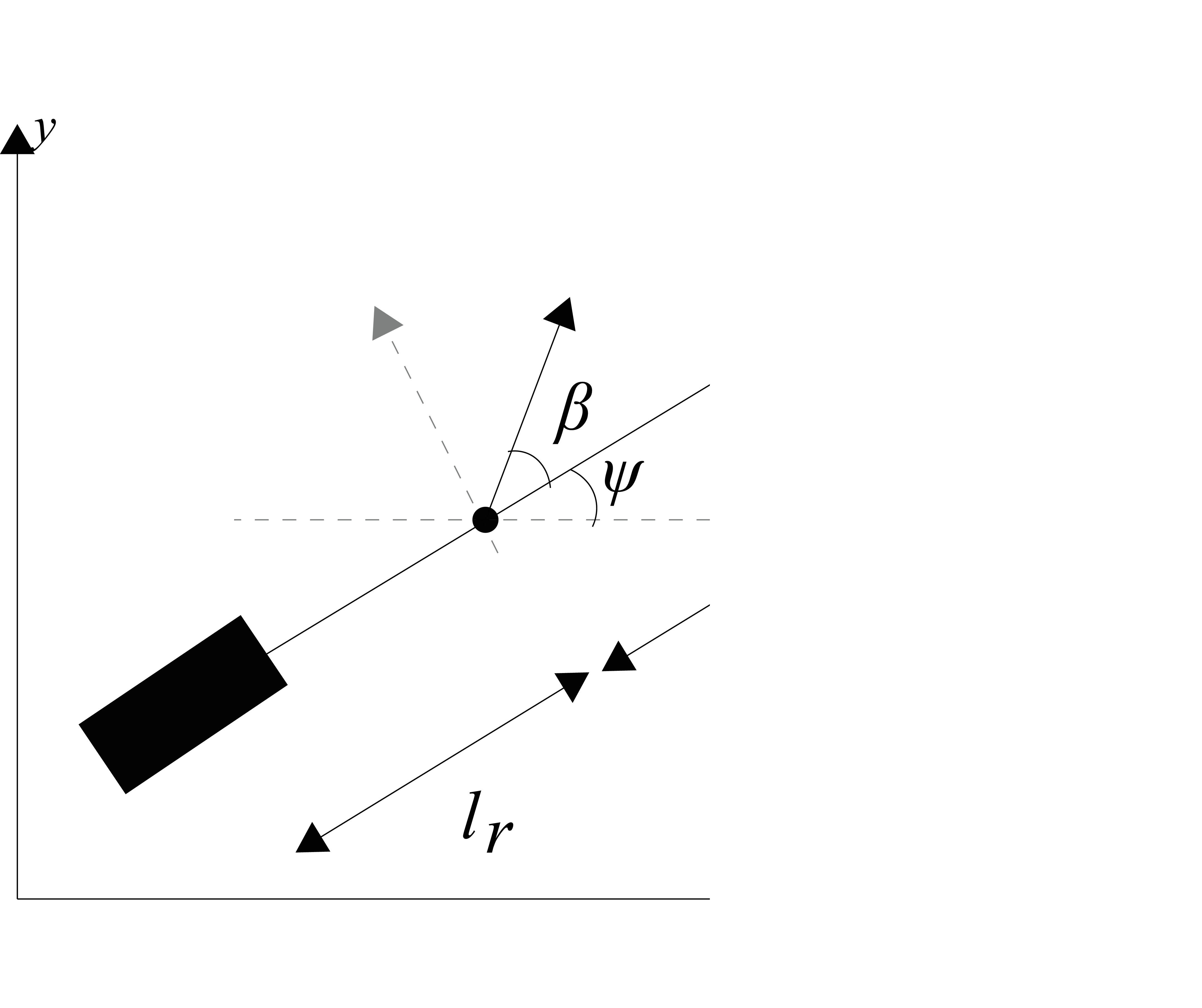}
        \caption{Bicycle model of simulated agent}
        \label{fig:minipage1}
    \end{minipage}\hspace{0.7 cm}
    \quad
    \begin{minipage}[b]{0.45\linewidth}
        \hspace*{-1 cm}
        \includegraphics[scale = 0.3]{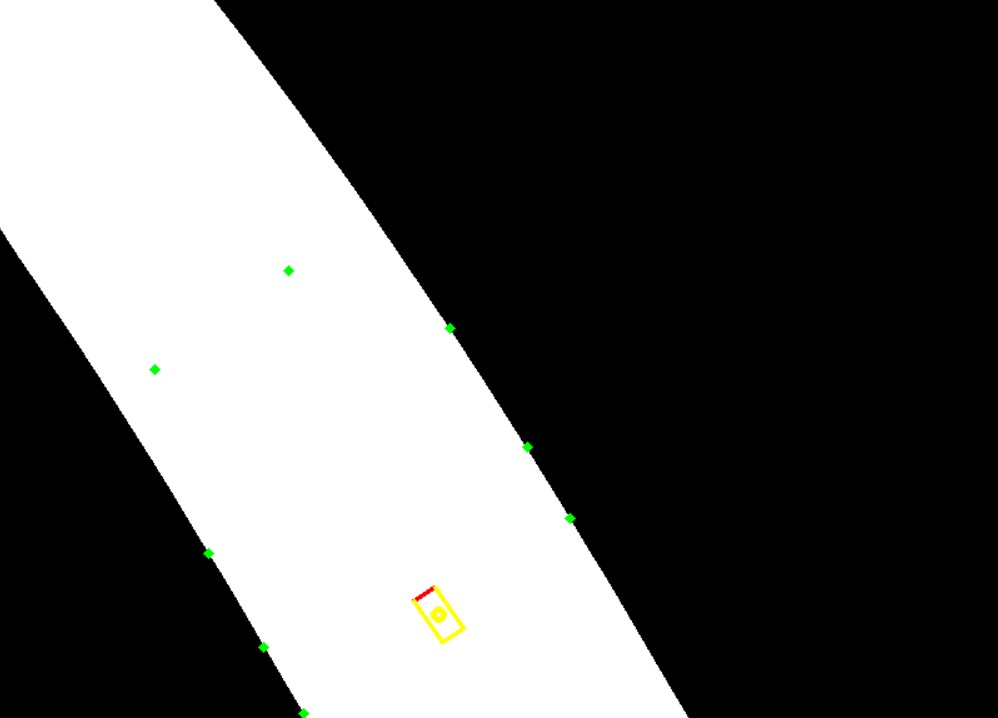}
        \caption{OGM Environment}
        \label{env}
    \end{minipage}
\end{figure}

\textbf{Agent Inputs} The green points in Fig \ref{env} represent the inputs to the simulation agent. These vision points effectively act as ranging sensors and aid the car in measuring the distance of the car from the boundary of the lane at discrete angles. As shown in the illustration, the vision points are offset to each other at an angle $\beta$.

\begin{figure}[h]
    \centering
    \begin{minipage}[b]{0.45\linewidth}
        \includegraphics[scale = 0.2]{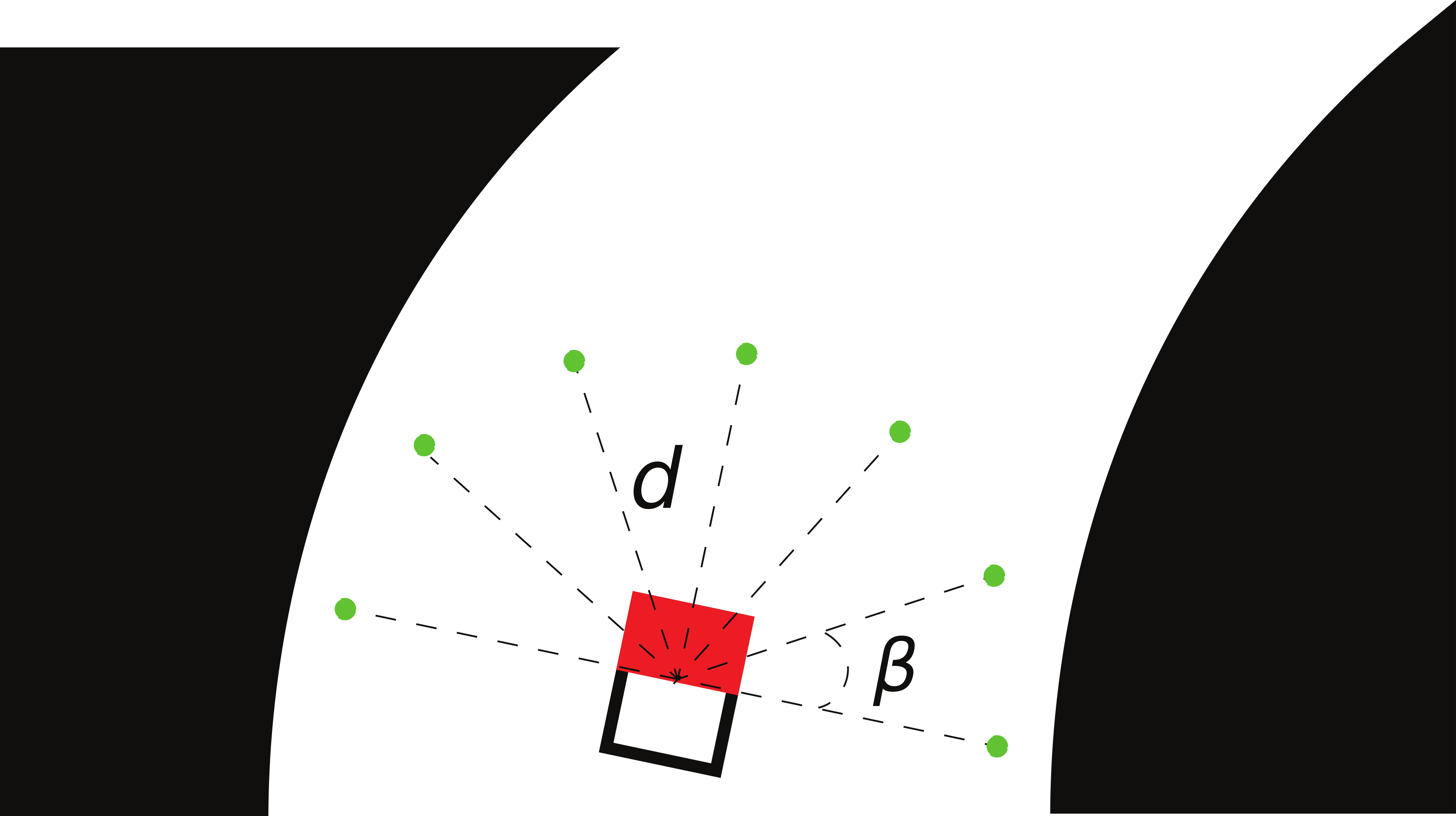}
        
    \end{minipage}\hspace{0.7 cm}
    \quad
    \begin{minipage}[b]{0.45\linewidth}
        \hspace*{-1 cm}
        \includegraphics[scale = 0.215]{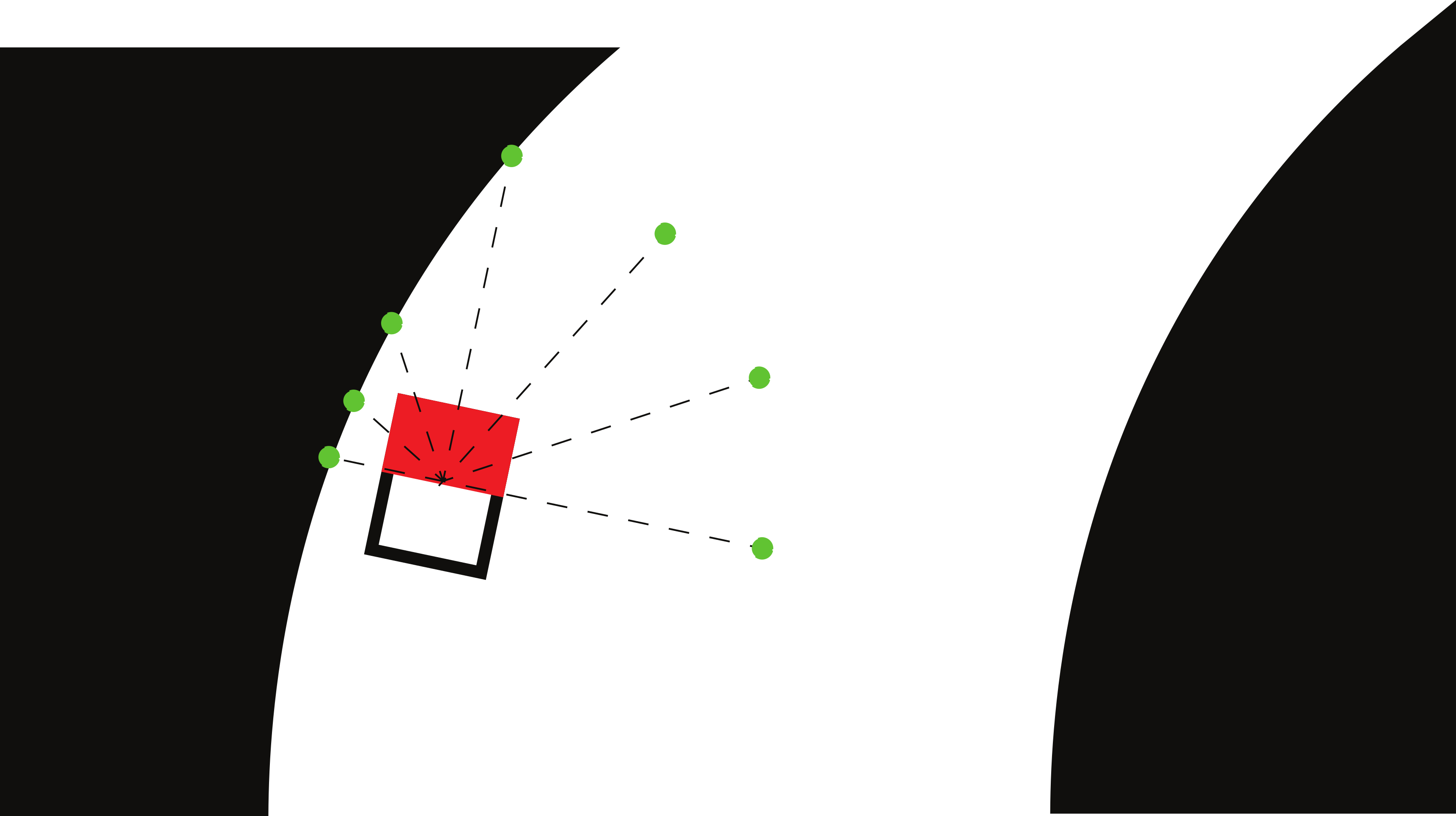}

    \end{minipage}
    
    \caption{Inputs to simulated agent - The agent has $\lambda$ ranging sensors that provide distances to the track. The agent learns to generate optimal high speed trajectories}
    
\end{figure}

At every instant, the car's state can be represented by the distances measured by these ranging sensors. Thus if we take $n_v$ vision points, the car's position can be represented as an array of $n_v$ distances $\{d_1, d_2, d_3, d_4 ... d_{n_v}\}$, where each of the $d_i$ is the distances of the vision points from the centre of mass of the car. These form the set of inputs to the neural network that acts on the simplified environment.

\textbf{Track Generation} : The tracks are procedurally generated with the following assumptions - (1) The width of the track is constant, (2) The boundary lanes follow a Bezier spline approximation. We represent the scene in a binarized fashion where the white region represents the lane, and the black region is anything that is not a lane.


\subsubsection{Genetic Algorithm}

We use a simple mutation strategy for training the agent for trajectory generation. Starting with $n$ initial spawns of the car, we let the agent run in the environment and retain only the top $m$ survivors. Then, at each generation, we mutate weights and biases using noise sampled from a Gaussian distribution. Using this strategy, the average fitness of the survivors increases over the number of generations, indicating that the agents learn to generate trajectories.

As seen in figure 4, the agent receives $\lambda$ distances that is given by the following equation:

\begin{equation}
    D_i = min(L, \theta _i)    
\end{equation}

where $L$ is a parameter that dictates the maximum distance till which the agent can see. We treat this value as a tunable hyperparameter and set it to 150px in our experiments.

The reward function used for the trajectory generation agent is defined as follows:
\begin{equation}
    R = \alpha v + \beta d
\end{equation}
where $\alpha$ and $\beta$ are tunable parameters, and $v$ and $d$ are the velocity and distance of the agent respectively.


\section{End to End Training : From RGB to Trajectory}

\subsection{Dataset Generation}

\begin{figure}
    \centering
    \includegraphics[width = \linewidth]{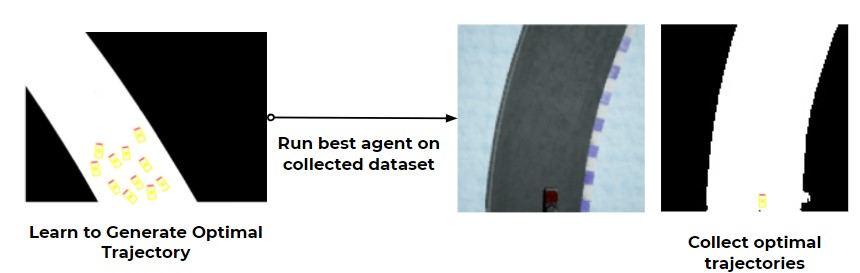}
    \caption{Generating optimal trajectories for a dataset}
    \label{fig:datasetgen}
\end{figure}
In order to obtain smooth trajectories, we remodel our problem by predicting control points of Bezier curves rather than directly predicting the waypoints, as proposed in \cite{weiss_bezier_nodate}. Hence, we represent our trajectories by approximating them with Bezier curves and sampling points from the Bezier curve at regular intervals of increasing y coordinate in the environment. We call this the \textit{trajectory embedding}, which would be the final prediction of the learning-based model.

We use the agent trained on the simplified simulation as an oracle that generates accurate and robust trajectories given a good occupancy map of the environment. To generate training data for the actual perception and trajectory generation block, we generate a dataset using CARLA by driving on the racetrack environment and collecting segmentation maps of the top view and the RGB front-facing camera. We use the obtained top view segmentation from CARLA and generate an OGM. 
Using the trained oracle agent, we run the agent on the OGM's to generate high-speed trajectories for the car to follow. Hence, our dataset consists of a three-way mapping between - (1) Front facing RGB, (2) OGM of environment, (3) High-speed trajectories generated from the OGM.

\subsection{Training}

We train the network (Fig \ref{fig:overall})in an end to end supervised manner using a summation of 4 losses - (1) Binary cross entropy loss for Auto-encoding, (2) Binary cross entropy for OGM prediction, (3) KL- Divergence loss for VAE, (4) MSE loss for trajectory prediction.
We downscale the images to 128x128px during training.

\begin{equation}
    \mathbb{L}_\textit{BCE, OGM} = \frac{1}{n}\sum _{\forall n}w_{n}\left[y'_{n} \cdot \log x'_{n}+\left(1-y'_{n}\right) \cdot \log \left(1-x^1_{n}\right)\right]
\end{equation}
\begin{equation}
    \mathbb{L}_\textit{BCE, AE} = \frac{1}{n}\sum _{\forall n}w_{n}\left[y''_{n} \cdot \log x''_{n}+\left(1-y''_{n}\right) \cdot \log \left(1-x''_{n}\right)\right]
\end{equation}
\begin{equation}
    \mathbb{L}_\textit{KLD} =  \frac{1}{n}\sum_{i = 1} ^k \sigma_{i}^{2}+\mu_{i}^{2}-\log \left(\sigma_{i}\right)-1 
\end{equation}
\begin{equation}
\mathbb{L}_\textit{MSE} = \frac{1}{k}\sum_{i = 1} ^k \left(x'''_{n}-y'''_{n}\right)^{2}
\end{equation}

Where $y'$ is the target pixel for the OGM, $y''$ is the target pixel for the auto-encoder, $y'''$ is the target trajectory point, $\sigma$ and $\mu$ are the standard deviation and means predicted from the VAE.
\section{Controller}
Once the trajectory is obtained, the points corresponding of trajectory are being tracked. The control of the car is divided into two steps; longitudinal control and lateral control. \\

For longitudinal control i.e. controlling the throttle to the car, we use PID controller to track a point on the trajectory. 
\begin{equation}
    a = K_pdy+K_d\frac{dy}{dt}+K_i\int ydt
\end{equation}
The gains $K_p$, $K_i$,$K_d$ are found by iterating through several values.The term $dy$ and $dt$ are actually discrete values and thus we replace $dy$ by ($y_{des} - y$) and $dt$ by $\Delta t$

For lateral control i.e. steering input to the car, we use a Stanley controller.
\begin{figure}[h]
    \centering
    \includegraphics[scale = 0.2]{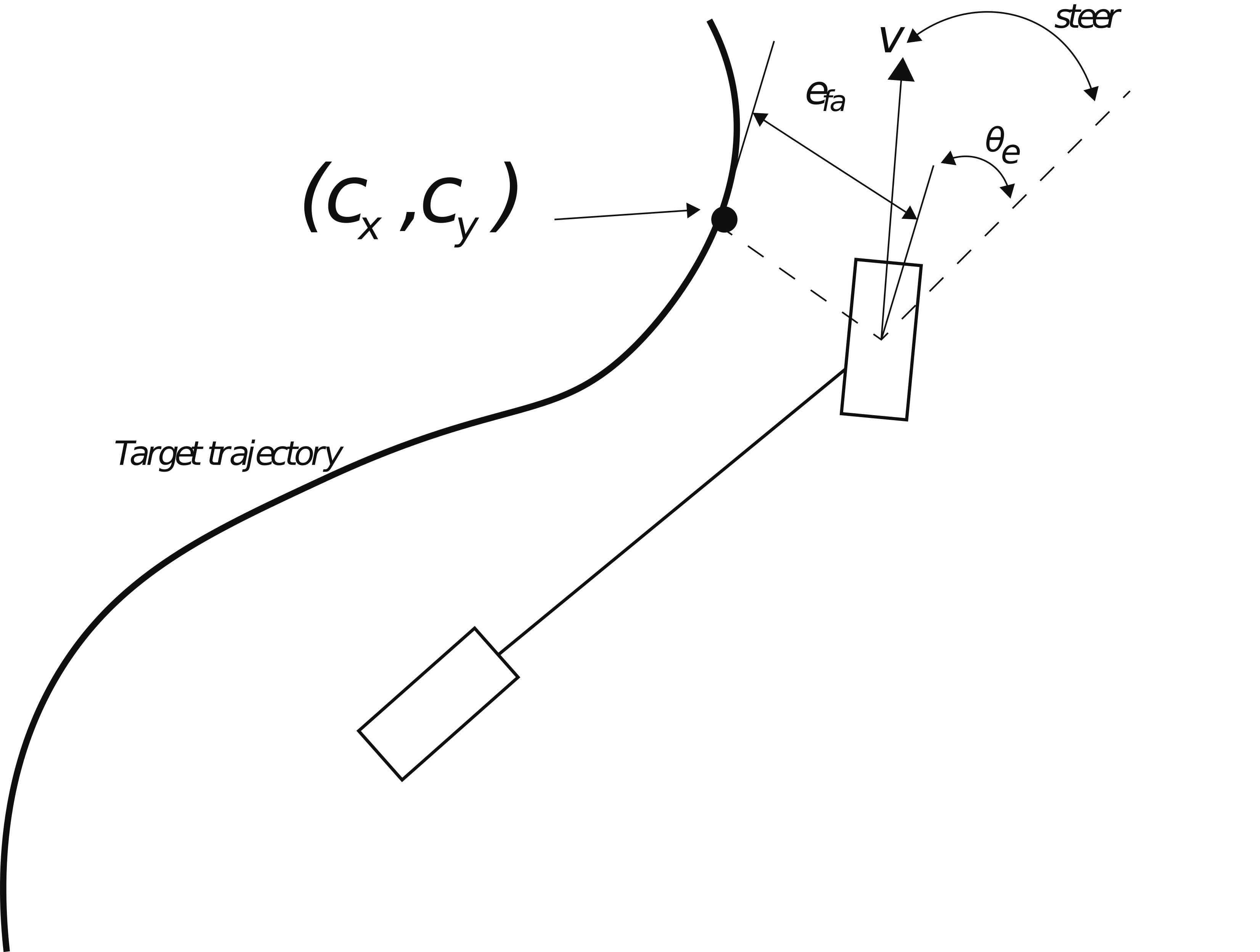}
    \caption{Geometry of a stanley controller}
    \label{stanley}
\end{figure}
We used the Stanley controller as it takes into consideration both the heading angle error and also the cross track error. The equations for a typical Stanley controller is given by:
\begin{equation}
        \theta_e = \theta - \theta_p
\end{equation}
\begin{equation}
        steer = \theta_e(t) + \tan^{-1}\big(\frac{Ke_{fa}(t)}{v_x(t)}\big)
\end{equation}
Here, $K$ is the gain parameter, $e_{fa}$ is the cross track error of the car from the tracking point and $v_x$ is the velocity of the car in the lateral direction. 

\begin{figure}
    \centering
    \includegraphics[width = \linewidth]{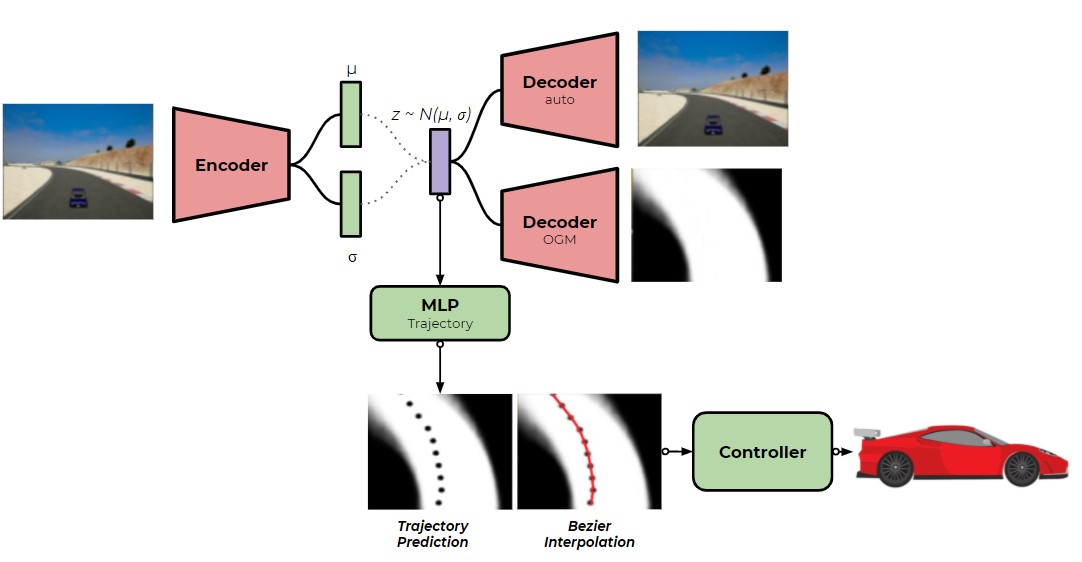}
    \caption{Overall Pipeline : We use a 3 layer MLP to predict trajectory embeddings from the latent space of the VAE.}
    \label{fig:overall}
\end{figure}

\section{Results}
\begin{figure}[h!]
    \centering
    \includegraphics[scale = 0.3]{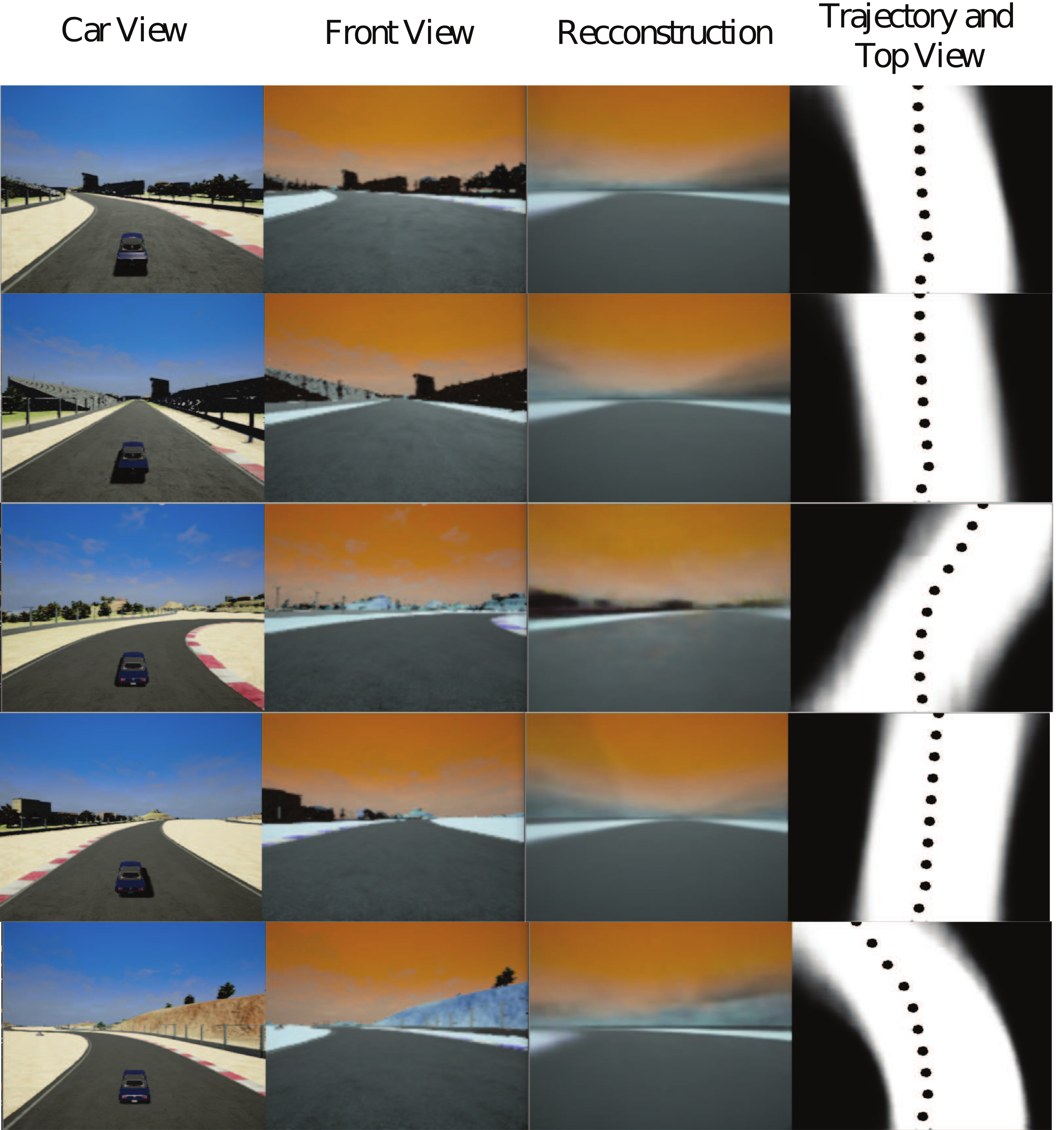}
    \caption{Predictions from the final End to End inference model. We map images directly to trajectories by learning intermediate latent representations.}
    \label{fig:predictions}
\end{figure}

The network was trained on $\mathtt{\sim} 18,000$  images taken from the CARLA racetrack environment, with trajectories generated by the oracle agent for each of these images. 

We achieved an average inference time of ~260 FPS on CARLA when run on a device with a GTX 1660Ti GPU, i7-9750H, and 16GB RAM. The obtained speed is signifcantly faster than what is required for realtime inference during racing.

\begin{table}[ht!]
    \centering
        \begin{tabular}{llll}\toprule
        {Method} & {Successful Laps} &{$t_{lap, avg}$ (s)} &{$t_{first failure}$  (s)} \\ \midrule
        {CNN Regression} & 0 & \textemdash & 30  \\
        {End to end}  & 5  & 1378 & \textemdash   \\
        \bottomrule
    \end{tabular}
    \caption{Experimental results (5 Lap benchmark). }
    \label{tab:results}
\end{table}

Table \ref{tab:results} show some quantitative results compared to a direct image to steering angle approach without throttle control. To train the regression based algorithm, we collected discretized expert steering data (left or right) and trained the network using mean squared error loss. We define a failure case as any condition that results in the car going off-track. We see that the CNN approach is significantly inferior compared to our method as it fails to complete laps successfully.  In our experiments, to maintain a fair comparison between the regression based approach and end to end approach, we maintain a constant car speed.

\section{Conclusion \& Future Work}

In this work, we present an end to end pipeline based on intermediate representations to generate trajectories from input images. We use a novel speciation based learning strategy to generate high quality racing trajectories using a simplified version of the environment. By combining robust deep learning approaches for perception and advances in model free learning for high speed racing, our framework robustly maps images to trajectories at high inference rates.

Implementation of our framework on a 1/10th scale testbed has begun, and our simulation experiments and models were designed keeping the hardware constraints of onboard computation in mind. 

In this work, we assumed a simplistic bicycle model of the agent for generating the trajectories from the occupancy grid. By changing the dynamics of the car to more closely match an actual car using better models, we can obtain trajectories which can potentially be of a much higher quality that our simple approximation. Moreover, it is possible to extend this method to dynamically changing scenes such as multi agent racing. However, that will require an increased dependence on temporal data, which can potentially be tackled by introducing recurrent memory units in both the trajectory generation stage (where obstacles are modelled as moving through the lane), and the autoencoding stage.


\nocite{*}
\bibliographystyle{acm}
\bibliography{main}  

\begin{thebibliography}{10}

\bibitem{ALCALA2020104270}
{\sc AlcalÃ¡, E., Puig, V., Quevedo, J., and Rosolia, U.}
\newblock Autonomous racing using linear parameter varying-model predictive
  control (lpv-mpc).
\newblock {\em Control Engineering Practice 95\/} (2020), 104270.

\bibitem{altche_lstm_2017}
{\sc Altche, F., and de~La~Fortelle, A.}
\newblock An {LSTM} network for highway trajectory prediction.
\newblock In {\em 2017 {IEEE} 20th International Conference on Intelligent
  Transportation Systems ({ITSC})}, {IEEE}, pp.~353--359.

\bibitem{amini_variational_2018}
{\sc Amini, A., Schwarting, W., Rosman, G., Araki, B., Karaman, S., and Rus,
  D.}
\newblock Variational autoencoder for end-to-end control of autonomous driving
  with novelty detection and training de-biasing.
\newblock In {\em 2018 {IEEE}/{RSJ} International Conference on Intelligent
  Robots and Systems ({IROS})}, {IEEE}, pp.~568--575.

\bibitem{DBLP:journals/corr/abs-2010-08776}
{\sc Bojarski, M., Chen, C., Daw, J., Degirmenci, A., Deri, J., Firner, B.,
  Flepp, B., Gogri, S., Hong, J., Jackel, L.~D., Jia, Z., Lee, B.~J., Liu, B.,
  Liu, F., Muller, U., Payne, S., Prasad, N. K.~N., Provodin, A., Roach, J.,
  Rvachov, T., Tadimeti, N., van Engelen, J.~E., Wen, H., Yang, E., and Yang,
  Z.}
\newblock The {NVIDIA} pilotnet experiments.
\newblock {\em CoRR abs/2010.08776\/} (2020).

\bibitem{airsim}
{\sc Bonatti, R., Madaan, R., Vineet, V., Scherer, S., and Kapoor, A.}
\newblock Learning visuomotor policies for aerial navigation using cross-modal
  representations.
\newblock In {\em 2020 IEEE/RSJ International Conference on Intelligent Robots
  and Systems (IROS)\/} (2020), pp.~1637--1644.

\bibitem{Dosovitskiy17}
{\sc Dosovitskiy, A., Ros, G., Codevilla, F., Lopez, A., and Koltun, V.}
\newblock {CARLA}: {An} open urban driving simulator.
\newblock In {\em Proceedings of the 1st Annual Conference on Robot Learning\/}
  (2017), pp.~1--16.

\bibitem{superhuman}
{\sc Fuchs, F., Song, Y., Kaufmann, E., Scaramuzza, D., and Dürr, P.}
\newblock Super-human performance in gran turismo sport using deep
  reinforcement learning.
\newblock {\em IEEE Robotics and Automation Letters 6}, 3 (2021), 4257--4264.

\bibitem{Gauss1857}
{\sc Gauss, C.~F., and Davis, C.~H.}
\newblock Theory of the motion of the heavenly bodies moving about the sun in
  conic sections.
\newblock {\em Gauss's Theoria Motus 76}, 1 (1857), 5--23.

\bibitem{8754713}
{\sc Kabzan, J., Hewing, L., Liniger, A., and Zeilinger, M.~N.}
\newblock Learning-based model predictive control for autonomous racing.
\newblock {\em IEEE Robotics and Automation Letters 4}, 4 (2019), 3363--3370.

\bibitem{lavalle_randomized_1999}
{\sc {LaValle}, S., and Kuffner, J.}
\newblock Randomized kinodynamic planning.
\newblock In {\em Proceedings 1999 {IEEE} International Conference on Robotics
  and Automation (Cat. No.99CH36288C)}, vol.~1, pp.~473--479 vol.1.
\newblock {ISSN}: 1050-4729.

\bibitem{li_sparse_2015}
{\sc Li, Y., Littlefield, Z., and Bekris, K.~E.}
\newblock Sparse methods for efficient asymptotically optimal kinodynamic
  planning.
\newblock In {\em Algorithmic Foundations of Robotics {XI}: Selected
  Contributions of the Eleventh International Workshop on the Algorithmic
  Foundations of Robotics}, H.~L. Akin, N.~M. Amato, V.~Isler, and A.~F.
  van~der Stappen, Eds., Springer Tracts in Advanced Robotics. Springer
  International Publishing, pp.~263--282.

\bibitem{morari15}
{\sc Liniger, A., Domahidi, A., and Morari, M.}
\newblock Optimization-based autonomous racing of 1:43 scale rc cars.
\newblock {\em Optimal Control Applications and Methods 36}, 5 (2015),
  628--647.

\bibitem{loukkal_driving_2021}
{\sc Loukkal, A., Grandvalet, Y., Drummond, T., and Li, Y.}
\newblock Driving among flatmobiles: Bird-eye-view occupancy grids from a
  monocular camera for holistic trajectory planning.
\newblock In {\em 2021 {IEEE} Winter Conference on Applications of Computer
  Vision ({WACV})}, {IEEE}, pp.~51--60.

\bibitem{flatmobiles}
{\sc Loukkal, A., Grandvalet, Y., Drummond, T., and Li, Y.}
\newblock Driving among flatmobiles: Bird-eye-view occupancy grids from a
  monocular camera for holistic trajectory planning.
\newblock In {\em 2021 {IEEE} Winter Conference on Applications of Computer
  Vision ({WACV})}, pp.~51--60.
\newblock {ISSN}: 2642-9381.

\bibitem{lienkamp2019}
{\sc Nobis, F., Betz, J., Hermansdorfer, L., and Lienkamp, M.}
\newblock Autonomous racing: A comparison of slam algorithms for large scale
  outdoor environments.
\newblock pp.~82--89.

\bibitem{pan_agile_2019}
{\sc Pan, Y., Cheng, C.-A., Saigol, K., Lee, K., Yan, X., Theodorou, E., and
  Boots, B.}
\newblock Agile autonomous driving using end-to-end deep imitation learning.

\bibitem{NEURIPS2019_9015}
{\sc Paszke, A., Gross, S., Massa, F., Lerer, A., Bradbury, J., Chanan, G.,
  Killeen, T., Lin, Z., Gimelshein, N., Antiga, L., Desmaison, A., Kopf, A.,
  Yang, E., DeVito, Z., Raison, M., Tejani, A., Chilamkurthy, S., Steiner, B.,
  Fang, L., Bai, J., and Chintala, S.}
\newblock Pytorch: An imperative style, high-performance deep learning library.
\newblock In {\em Advances in Neural Information Processing Systems 32},
  H.~Wallach, H.~Larochelle, A.~Beygelzimer, F.~d\textquotesingle
  Alch\'{e}-Buc, E.~Fox, and R.~Garnett, Eds. Curran Associates, Inc., 2019,
  pp.~8024--8035.

\bibitem{perot_end--end_2017}
{\sc Perot, E., Jaritz, M., Toromanoff, M., and De~Charette, R.}
\newblock End-to-end driving in a realistic racing game with deep reinforcement
  learning.
\newblock In {\em 2017 {IEEE} Conference on Computer Vision and Pattern
  Recognition Workshops ({CVPRW})}, pp.~474--475.
\newblock {ISSN}: 2160-7516.

\bibitem{e2erl}
{\sc Perot, E., Jaritz, M., Toromanoff, M., and De~Charette, R.}
\newblock End-to-end driving in a realistic racing game with deep reinforcement
  learning.
\newblock In {\em 2017 IEEE Conference on Computer Vision and Pattern
  Recognition Workshops (CVPRW)\/} (2017), pp.~474--475.

\bibitem{7963748}
{\sc Rosolia, U., Carvalho, A., and Borrelli, F.}
\newblock Autonomous racing using learning model predictive control.
\newblock In {\em 2017 American Control Conference (ACC)\/} (2017),
  pp.~5115--5120.

\bibitem{TALVALA2011137}
{\sc Talvala, K.~L., Kritayakirana, K., and Gerdes, J.~C.}
\newblock Pushing the limits: From lanekeeping to autonomous racing.
\newblock {\em Annual Reviews in Control 35}, 1 (2011), 137--148.

\bibitem{weiss_bezier_nodate}
{\sc Weiss, T., Babu, V.~S., and Behl, M.}
\newblock Be`zier curve based end-to-end trajectory synthesis for agile
  autonomous driving.
\newblock 10.

\bibitem{9116486}
{\sc Weiss, T., and Behl, M.}
\newblock Deepracing: A framework for autonomous racing.
\newblock In {\em 2020 Design, Automation Test in Europe Conference Exhibition
  (DATE)\/} (2020), pp.~1163--1168.

\bibitem{https://doi.org/10.1002/atr.1359}
{\sc Yang, I., Kim, H.~J., Jeon, W.~H., and Kim, H.}
\newblock Development of realistic shortest path algorithm considering lane
  changes.
\newblock {\em Journal of Advanced Transportation 50}, 4 (2016), 541--551.

\end{thebibliography}

\end{document}